\documentclass[journal]{IEEEtran}
\usepackage{amsmath}
\usepackage{indentfirst}
\usepackage{booktabs}
\usepackage{subfigure}
\usepackage{epsfig}
\usepackage{amssymb}
\usepackage{graphicx}
\usepackage{float}
\usepackage{setspace}
\usepackage{tabularx,booktabs}
\usepackage{url}
\usepackage{amsfonts,amssymb}
\usepackage{dsfont}
\usepackage{enumerate}
\usepackage{algorithmic}
\usepackage[linesnumbered,ruled]{algorithm2e}
\usepackage[numbers,sort&compress]{natbib}
\usepackage{multirow}
\usepackage{rotating}
\usepackage{sverb, longtable}
\ifCLASSINFOpdf
\else
\fi
\begin{document}
%
\title{Generalization of Dempster--Shafer theory: A complex belief function}
%
%
%
%

\author{Fuyuan Xiao
\IEEEcompsocitemizethanks{\IEEEcompsocthanksitem F. Xiao is with the School of Computer and Information Science, Southwest University, No.2 Tiansheng Road, BeiBei District, Chongqing, 400715, China.\protect\\
E-mail: xiaofuyuan@swu.edu.cn
}
}

%
%

\markboth{}%
{
}
%



\IEEEtitleabstractindextext{%
\begin{abstract}
Dempster--Shafer evidence theory has been widely used in various fields of applications, because of the flexibility and effectiveness in modeling uncertainties without prior information.
However, the existing evidence theory is insufficient to consider the situations where it has no capability to express the fluctuations of data at a given phase of time during their execution, and the uncertainty and imprecision which are inevitably involved in the data occur concurrently with changes to the phase or periodicity of the data.
In this paper, therefore, a generalized Dempster--Shafer evidence theory is proposed.
To be specific, a mass function in the generalized Dempster--Shafer evidence theory is modeled by a complex number, called as a complex basic belief assignment, which has more powerful ability to express uncertain information.
Based on that, a generalized Dempster's combination rule is exploited.
In contrast to the classical Dempster's combination rule, the condition in terms of the conflict coefficient between the evidences $\mathds{K} < 1$ is released in the generalized Dempster's combination rule.
Hence, it is more general and applicable than the classical Dempster's combination rule.
When the complex mass function is degenerated from complex numbers to real numbers, the generalized Dempster's combination rule degenerates to the classical evidence theory under the condition that the conflict coefficient between the evidences $\mathds{K}$ is less than 1.
In a word, this generalized Dempster--Shafer evidence theory provides a promising way to model and handle more uncertain information.
\end{abstract}

}

\maketitle

\IEEEdisplaynontitleabstractindextext

%
\IEEEpeerreviewmaketitle

\section{Introduction}\label{Introduction}
How to measure the uncertainty has been an attracting issue in a variety of areas~\cite{yager2018using,zavadskas2014extension,fu2018data}.
The amount of theories had been presented and developed for measuring the uncertainty, including
the extended fuzzy sets~\cite{fei2019interval,Xiao2019Divergence},
fuzzy soft sets~\cite{feng2016soft},
evidence theory~\cite{Sun2019A,jiang2018Correlation},
D~numbers theory~\cite{Zhao2019Dnumbers,DNTIJAR2019},
Z~numbers~\cite{Jiang2019Znetwork,kang2019environmental},
R~numbers~\cite{Seiti2019Rnumbers,seiti2019developing},
entropy-based~\cite{cao2018Inherent,wang2018analysis},
and information quality~\cite{yager2019using}.
These theories were broadly applied in various fields, such as the
selection~\cite{seiti2018extending,deng2019evaluating},
recognition~\cite{Geng2019Saliency},
prediction~\cite{zhou2019robust},
medical diagnosis~\cite{cao2019extraction},
and decision-making~\cite{Xiao2018Anovelmulti,feng2018another,Xiao2019Amultiplecriteria}.

As one of the most effective tools of uncertainty reasoning,
Dempster--Shafer (DS) evidence theory~\cite{Dempster1967Upper,shafer1976mathematical} can model the uncertainty without prior information in a flexible and effective manner~\cite{XDeng2019Polymatrix,suxiaoyan2019,yager2017soft}.
The fusion results generated by Dempster's combination rule are fault-tolerant which can be more sufficient and accurate to support the decision-making~\cite{lilusu2018,yager2018satisfying}, while the uncertainty can be characterized quantitatively and further be reduced in the process of combination~\cite{seiti2018risk,Xiao2018AHybridFuzzy,yager2019generalized}.
Besides, the Dempster-Shafer theory satisfies the commutative and associative laws, so that it has been extensively applied in various fields~\cite{Jiang2018Information,yager2018fuzzy}.
Nevertheless, through carefully studying the existing methods of evidence theory, it is found that none of these models have the capability to express the fluctuations of data at a given phase of time during their execution.
Furthermore, in daily life, uncertainty and imprecision which are inevitably involved in the data occur concurrently with changes to the phase or periodicity of the data.
As a result, the existing evidence theories are insufficient to consider these kinds of information, so that some information would loss during the model and process of data.

In this paper, therefore, a generalized Dempster--Shafer (GDS) evidence theory is proposed.
To be specific, a mass function in the GDS evidence theory is modeled by a complex number, called as a complex mass function, which has more powerful ability to express uncertain information.
On this basis, a generalized Dempster's combination rule is exploited.
Compared with the traditional Dempster's combination rule, the condition in terms of the conflict coefficient between two evidences $\mathds{K} < 1$ is released in the generalized Dempster's combination rule.
Hence, the proposed method is more general and applicable than the traditional Dempster's combination rule.
In particular, when the complex mass function is degenerated from complex numbers to real numbers, the generalized Dempster's combination rule degenerates to the traditional evidence theory under the condition that the conflict coefficient between two evidences $\mathds{K}$ is less than 1.
In this context, the GDS evidence theory
provides a new framework to be more capable of modeling and handling the uncertainty.
Meanwhile, several numerical examples are provided to illustrate the feasibility of the GDS evidence theory.
Additionally, an algorithm for decision-making is devised based on the GDS evidence theory.
Finally, an application of the new algorithm is implemented to solve the medical diagnosis problem.
The results validate the practicability and effectiveness of the proposed algorithm.

The rest of this paper is organised as follows.
The preliminaries, including complex number and Dempster--Shafer evidence theory are briefly introduced in Section~\ref{Preliminaries}.
The new GDS evidence theory is proposed in Section~\ref{Proposed method}.
Section~\ref{Experiments} provides numerical examples to illustrate the feasibility of the GDS evidence theory.
Finally, the conclusion is given in Section~\ref{Conclusion}.

\section{Preliminaries}\label{Preliminaries}

\subsection{Complex number~\cite{ablowitz2003complex}}\label{Complexnumber}
A complex number $z$ is defined as an ordered pair of real numbers
\begin{equation}\label{eq_complexnumber}
z = x + yi,
\end{equation}
where $x$ and $y$ are real numbers and $i$ is the imaginary unit, satisfying $i^2 = -1$.
This is called the ``rectangular'' form or ``Cartesian'' form.

It can also expressed in polar form, denoted by
\begin{equation}\label{eq_complexnumber}
z=r e^{i \theta},
\end{equation}
where $r > 0$ represents the modulus or magnitude of the complex number $z$ and $\theta$ represents the angle or phase of the complex number $z$.

By using the Euler's relation,
\begin{equation}\label{eq_Euler'srelation}
e^{i \theta} = \cos(\theta) +i \sin(\theta),
\end{equation}
the modulus or magnitude and angle or phase of the complex number can be expressed as
\begin{equation}\label{eq_magnitudeandphase}
r=\sqrt{x^2+y^2}, \text{ and } \theta = \arctan(\frac{y}{x}) = \tan^{-1}(\frac{y}{x}),
\end{equation}
where $x = r \cos(\theta)$ and $y = r \sin(\theta)$.

The square of the absolute value is defined by
\begin{equation}\label{eq_squaremagnitude}
|z|^2=z\bar{z}=x^2+y^2,
\end{equation}
where $\bar{z}$ is the complex conjugate of $z$, i.e., $\bar{z}=x - yi$.

These relationships can be then obtained as
\begin{equation}\label{eq_relationship}
r=|z|, \text{ and } \theta = \angle z,
\end{equation}
where if $z$ is a real number (i.e., $y = 0$), then $r = |x|$.

The arithmetic of complex numbers is defined as follows:

Give two complex numbers $z_1=x_1 + y_1i$ and $z_2=x_2 + y_2i$, the addition is defined by
\begin{equation}\label{eq_addition}
z_1+z_2=(x_1 + y_1i)+(x_2 + y_2i)=(x_1+x_2)+(y_1+y_2)i.
\end{equation}

The subtraction is defined by
\begin{equation}\label{eq_subtraction}
z_1-z_2=(x_1 + y_1i)-(x_2 + y_2i)=(x_1-x_2)+(y_1-y_2)i.
\end{equation}

The multiplication is defined by
\begin{equation}\label{eq_multiplication}
(x_1 + y_1i)(x_2 + y_2i)=(x_1x_2-y_1y_2)+(x_1y_2+x_2y_1)i.
\end{equation}

The division is defined by
\begin{equation}\label{eq_division}
\frac{x_1 + y_1i}{x_2 + y_2i}=\frac{x_1x_2+y_1y_2}{x_2^2+y_2^2}+\frac{x_2y_1-x_1y_2}{x_2^2+y_2^2}i.
\end{equation}

\subsection{Dempster--Shafer evidence theory~\cite{Dempster1967Upper,shafer1976mathematical}}
Uncertain information is inevitable in practical applications~\cite{zavadskas2017model,zhou2018evidential,de2018robust}.
To handle the uncertainty problems in the process of information fusion, many integrated methods have been presented in recent years~\cite{yager2018multi,feng2019lexicographic,Wang2018uncertainty}, in which Dempster--Shafer (DS) evidence theory is very common used in the real applications~\cite{Li2018Generalized,liu2018classifier,gong2018Research}.
The basic concepts and definitions are described as below.

\newtheorem{myDef}{Definition}
\begin{myDef}(Frame of discernment)

Let $\Omega$ be a set of mutually exclusive and collective non-empty events, defined by
\begin{equation}\label{eq_Frameofdiscernment1}
 \Omega = \{F_{1}, F_{2}, \ldots, F_{i}, \ldots, F_{N}\},
\end{equation}
where $\Omega$ is a frame of discernment~\cite{Jiang2019IJIS}.

The power set of $\Omega$ is denoted as $2^{\Omega}$,
\begin{equation}\label{eq_Frameofdiscernment2}
\begin{aligned}
 2^{\Omega} = \{\emptyset, \{F_{1}\}, \{F_{2}\}, \ldots, \{F_{N}\}, \{F_{1}, F_{2}\}, \ldots, \{F_{1}, \\
 F_{2}, \ldots, F_{i}\}, \ldots, \Omega\},
\end{aligned}
\end{equation}
where $\emptyset$ represents an empty set.

If $A \in 2^{\Omega}$, $A$ is called a proposition~\cite{Xiao2019Multisensor}.
\end{myDef}

\begin{myDef}(Mass function)

A mass function $m$ in the frame of discernment $\Omega$ can be described as a mapping from $2^{\Omega}$ to [0, 1], defined as
\begin{equation}\label{eq_Massfunction1}
 m: \quad 2^{\Omega} \rightarrow [0, 1],
\end{equation}
satisfying the following conditions,
\begin{equation}
\label{eq_Massfunction2}
\begin{aligned}
 m(\emptyset) &= 0, \text{ and }
 \sum\limits_{A \in 2^{\Omega}} m(A) &= 1.
 \end{aligned}
\end{equation}
\end{myDef}

In the DS evidence theory, $m$ can also be called a basic belief assignment (BBA).
If $m(A)$ is greater than zero, where $A \in 2^{\Omega}$, $A$ is called a focal element.
The value of $m(A)$ represents how strongly the evidence supports the proposition $A$~\cite{Zhang2018DEMATEL,Sun2019GBPA}.

\begin{myDef}(Belief function)

Let $A$ be a proposition in the frame of discernment $\Omega$.
The belief function of proposition $A$, denoted as $Bel(A)$ is defined by
\begin{equation}\label{eq_belieffunction}
\begin{aligned}
Bel(A) = \sum\limits_{B \subseteq A} m(B).
\end{aligned}
\end{equation}
\end{myDef}

\begin{myDef}(Plausibility function)

Let $A$ be a proposition in the frame of discernment $\Omega$.
The plausibility function of proposition $A$, denoted as $Pl(A)$ is defined by
\begin{equation}\label{eq_plausibilityfunction}
\begin{aligned}
Pl(A) = \sum\limits_{B \cap A \neq \emptyset} m(B).
\end{aligned}
\end{equation}
\end{myDef}

The belief function $Bel(A)$ and plausibility function $Pl(A)$ represent the lower and upper bound functions of the proposition $A$, respectively~\cite{dezert2018total,Jiang2018KBS,Li2019TDBF}.
The value of $m(A)$ represents how strongly the evidence supports the proposition $A$~\cite{yager2019entailment}.
Various operations on the BBA are presented, like
negation~\cite{Gao2019negation,Gao2019generalizationnegation},
belief interval~\cite{han2016belief,Songyf2016},
divergence~\cite{Song2019divergence},
and entropy function~\cite{cui2019improved,yager2008entropy,dong2019combination}.

\begin{myDef}(Dempster's rule of combination)

Let $m_1$ and $m_2$ be two independent basic belief assignments (BBAs) in the frame of discernment $\Omega$.
The Dempster's rule of combination, denoted as $m = m_1 \oplus m_2$ is defined by
\begin{equation}\label{eq_Dempsterrule1}
{
m(C) = \left\{ \begin{array}{l}
\begin{array}{*{20}{c}}
{\frac{1}{1-K} \sum\limits_{A \cap B = C} m_1(A) m_2(B),}&{{\kern 30pt} C \neq \emptyset,}
\end{array}\\
\begin{array}{*{20}{c}}
{0,}&{{\kern 136pt} C = \emptyset,}
\end{array}
\end{array} \right.
}\end{equation}
with
\begin{equation}\label{eq_Dempsterrule2}
{
K = \sum\limits_{A \cap B = \emptyset} m_1(A) m_2(B),
}\end{equation}
where $A, B \in 2^{\Omega}$ 
and $K$ is the conflict coefficient between $m_1$ and $m_2$.
\end{myDef}

Notice that the Dempster's combination rule is only feasible under the situation where the conflict coefficient $K < 1$ for $m_1$ and $m_2$~\cite{liu2018combination,Zhanghp2018AIME}.
As an useful uncertainty processing methodology~\cite{yager2017maxitive,Xu2019DEMATEL,Jiang2019Uncertainty}, DS evidence theory was widely applied in various areas, like
reasoning~\cite{zhou2019evidential,fu2019multiple},
reliability evaluation~\cite{song2018sensor,fan2018evidence},
fault diagnosis~\cite{Zhanghp2019Weighted},
decision-making~\cite{zhou2017evidential,dezert2016decision},
and classification~\cite{liu2019evidence,liu2019new}.

%

\section{Generalized Dempster--Shafer evidence theory}\label{Proposed method}

Let $\Omega$ be a set of mutually exclusive and collective non-empty events, defined by
\begin{equation}\label{eq_Frameofdiscernment1}
 \Omega = \{E_{1}, E_{2}, \ldots, E_{i}, \ldots, E_{N}\},
\end{equation}
where $\Omega$ represents a frame of discernment.

The power set of $\Omega$ is denoted by $2^{\Omega}$, in which
\begin{equation}\label{eq_Frameofdiscernment2}
\begin{aligned}
 2^{\Omega} = \{\emptyset, \{E_{1}\}, \{E_{2}\}, \ldots, \{E_{N}\}, \{E_{1}, E_{2}\}, \ldots, \{E_{1}, \\
 E_{2}, \ldots, E_{i}\}, \ldots, \Omega\},
\end{aligned}
\end{equation}
and $\emptyset$ is an empty set.

\begin{myDef}(Complex mass function)\label{def_Complexmassfunction}

A complex mass function $\mathds{M}$ in the frame of discernment $\Omega$ is modeled as a complex number, which is represented as a mapping from $2^{\Omega}$ to $\mathbb{C}$, defined by
\begin{equation}\label{eq_GMassfunction1}
 \mathds{M}: \quad 2^{\Omega} \rightarrow \mathbb{C},
\end{equation}
satisfying the following conditions,
\begin{equation}\label{eq_Massfunction2}
\begin{aligned}
 &\mathds{M}(\emptyset) = 0, \\
 &\mathds{M}(A) = \mathbf{m}(A) e^{i \theta(A)}, \quad A \in 2^{\Omega} \\
 &\sum\limits_{A \in 2^{\Omega}} \mathds{M}(A) = 1,
\end{aligned}
\end{equation}
\end{myDef}
where $i = \sqrt{-1}$; $\mathbf{m}(A) \in [0, 1]$ representing the magnitude of the complex mass function $\mathds{M}(A)$;
$\theta(A) \in [-\pi, \pi]$ denoting a phase term.

In Eq.~(\ref{eq_Massfunction2}), $\mathds{M}(A)$ can also expressed in the ``rectangular'' form or ``Cartesian'' form, denoted by
\begin{equation}\label{eq_Massfunction3}
\mathds{M}(A) = x + yi, \quad A \in 2^{\Omega}
\end{equation}
with
\begin{equation}\label{eq_Massfunction4}
\sqrt{x^2+y^2} \in [0, 1].
\end{equation}

By using the Euler's relation, the magnitude and phase of the complex mass function $\mathds{M}(A)$ can be expressed as
\begin{equation}\label{eq_magnitudeandphase}
\mathbf{m}(A)=\sqrt{x^2+y^2}, \text{ and } \theta(A) = \arctan(\frac{y}{x}),
\end{equation}
where $x = \mathbf{m}(A) \cos(\theta(A))$ and $y = \mathbf{m}(A) \sin(\theta(A))$.

The square of the absolute value for $\mathds{M}(A)$ is defined by
\begin{equation}\label{eq_squaremagnitude}
|\mathds{M}(A)|^2=\mathds{M}(A)\mathds{\bar{M}}(A)=x^2+y^2,
\end{equation}
where $\mathds{\bar{M}}(A)$ is the complex conjugate of $\mathds{M}(A)$, such that $\mathds{\bar{M}}(A)=x - yi$.

These relationships can be then obtained as
\begin{equation}\label{eq_relationship}
\mathbf{m}(A)=|\mathds{M}(A)|, \text{ and } \theta(A) = \angle \mathds{M}(A),
\end{equation}
where if $\mathds{M}(A)$ is a real number (i.e., $y = 0$), then $\mathbf{m}(A) = |x|$.

The complex mass function $\mathds{M}$ modeled as a complex number in the generalized Dempster--Shafer (GDS) evidence theory can also be called a complex basic belief assignment (CBBA).

If $|\mathds{M}(A)|$ is greater than zero, where $A \in 2^{\Omega}$, $A$ is called a focal element of the complex mass function.
The value of $|\mathds{M}(A)|$ represents how strongly the evidence supports the proposition $A$.

\begin{myDef}(Complex belief function)

Let $A$ be a proposition in the frame of discernment $\Omega$.
The complex belief function of proposition $A$, denoted as $Bel_c(A)$ is defined by
\begin{equation}\label{eq_Gbelieffunction}
\begin{aligned}
Bel_c(A) = \sum\limits_{B \subseteq A} |\mathds{M}(B)|,
\end{aligned}
\end{equation}
where $|\mathds{M}(B)|$ represents the absolute value of $\mathds{M}(B)$.
\end{myDef}

\begin{myDef}(Complex plausibility function)

Let $A$ be a proposition in the frame of discernment $\Omega$.
The complex plausibility function of proposition $A$, denoted as $Pl_c(A)$ is defined by
\begin{equation}\label{eq_Gplausibilityfunction}
\begin{aligned}
Pl_c(A) = \sum\limits_{B \cap A \neq \emptyset} |\mathds{M}(B)|,
\end{aligned}
\end{equation}
where $|\mathds{M}(B)|$ represents the absolute value of $\mathds{M}(B)$.
\end{myDef}

Obviously, we can notice that $Pl_c(A) \geq Bel_c(A)$, in which the complex belief function $Bel_c(A)$ is the lower bound function of proposition $A$, and the complex plausibility function $Pl_c(A)$ is the upper bound function of proposition $A$.

%

\begin{myDef}(Generalized Dempster's rule of combination)

Let $\mathds{M}_1$ and $\mathds{M}_2$ be two independently complex basic belief assignments (CBBAs) in the frame of discernment $\Omega$.
The generalized Dempster's rule of combination, defined by $\mathds{M} = \mathds{M}_1 \oplus \mathds{M}_2$, which is called the orthogonal sum, is represented as below
\begin{equation}\label{eq_GDempsterrule1}
{
\mathds{M}(C) = \left\{ \begin{array}{l}
\begin{array}{*{20}{c}}
{\frac{1}{1-\mathds{K}} \sum\limits_{A \cap B = C} \mathds{M}_1(A) \mathds{M}_2(B),}&{{\kern 1pt} C \neq \emptyset,}
\end{array}\\
\begin{array}{*{20}{c}}
{0,}&{{\kern 109pt} C = \emptyset,}
\end{array}
\end{array} \right.
}\end{equation}
with
\begin{equation}\label{eq_GDempsterrule2}
{
\mathds{K} = \sum\limits_{A \cap B = \emptyset} \mathds{M}_1(A) \mathds{M}_2(B),
}\end{equation}
where $A, B \in 2^{\Omega}$ and $\mathds{K}$ is the conflict coefficient between the CBBAs $\mathds{M}_1$ and $\mathds{M}_2$.
\end{myDef}

\newtheorem{Remark}{Remark}
\begin{Remark}
The generalized Dempster's combination rule is only feasible under the situation where the conflict coefficient $\mathds{K} \neq 1$ for $\mathds{M}_1$ and $\mathds{M}_2$.
\end{Remark}

\begin{Remark}
Compared with the traditional Dempster's combination rule, the condition in terms of the conflict coefficient $\mathds{K} < 1$ is released in the generalized Dempster's combination rule so that it is more general and applicable than the traditional Dempster's combination rule.
\end{Remark}

\begin{Remark}
When the complex mass function is degenerated from complex numbers to real numbers, the generalized Dempster's combination rule degenerates to the traditional evidence theory under the condition that the conflict coefficient $\mathds{K} < 1$.
\end{Remark}

An example is given to illustrate that the condition $\mathds{K} < 1$ is released in the generalized Dempster's combination rule, where the variation of the magnitude of conflict coefficient $|\mathds{K}|$ between two CBBAs is depicted.
Note that $|\mathds{K}|$ can be calculated based on Eq.~(\ref{eq_relationship}).

\newtheorem{exmp}{Example}
\begin{exmp}\label{exa_conflictcoefficient}
\rm Supposing that there are two CBBAs $\mathds{M}_1$ and $\mathds{M}_2$ in the frame of discernment $\Omega=\{A,B\}$, and the two CBBAs are given as follows:
\end{exmp}
\begin{align*}
\mathds{M}_1:
&\mathds{M}_1(A)=\sqrt{x^2+y^2} e^{i \arctan (\frac{y}{x})}, \\
&\mathds{M}_1(B)=\sqrt{(1-x)^2+(-y)^2} e^{i \arctan (\frac{-y}{1-x})};\\
\mathds{M}_2:
&\mathds{M}_2(A)=\sqrt{0.5^2+0.5^2} e^{i \arctan (\frac{0.5}{0.5})}, \\
&\mathds{M}_2(B)=\sqrt{0.5^2+(-0.5)^2} e^{i \arctan (\frac{-0.5}{0.5})}.
\end{align*}

\begin{figure}[htpb]
\centering
\includegraphics[width=9cm,clip]{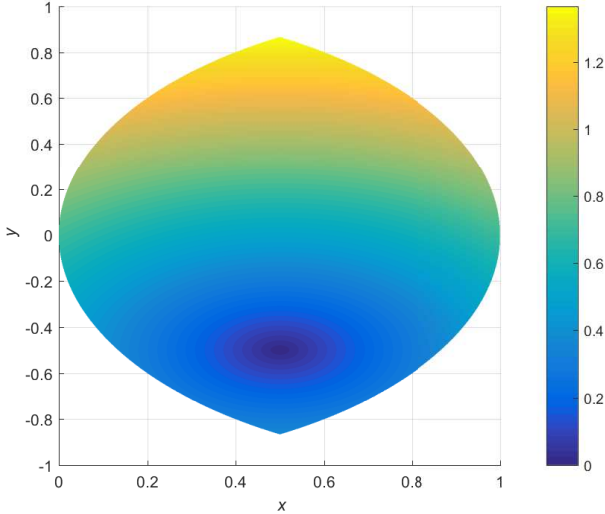}
 \caption{The variations of parameters $x$ and $y$.}\label{variationsofparameters}
\end{figure}

Since $|\mathds{M}_1(A)| = \sqrt{x^2+y^2}$ and $|\mathds{M}_1(B)| = \sqrt{(1-x)^2+(-y)^2}$, according to Definition~\ref{def_Complexmassfunction}, the parameters $x$ is set within $[0,1]$ and $y$ is set within $[-1,1]$ satisfying the conditions that $\sqrt{x^2+y^2} \in [0,1]$ and $\sqrt{(1-x)^2+(-y)^2} \in [0,1]$ at the same time, where the variations of parameters $x$ and $y$ are shown in Fig.~\ref{variationsofparameters}.

\begin{figure*}[htbp]
\centering
\subfigure[]{
\label{conflictcoefficient:a} 
\includegraphics[width=.45\linewidth,clip]{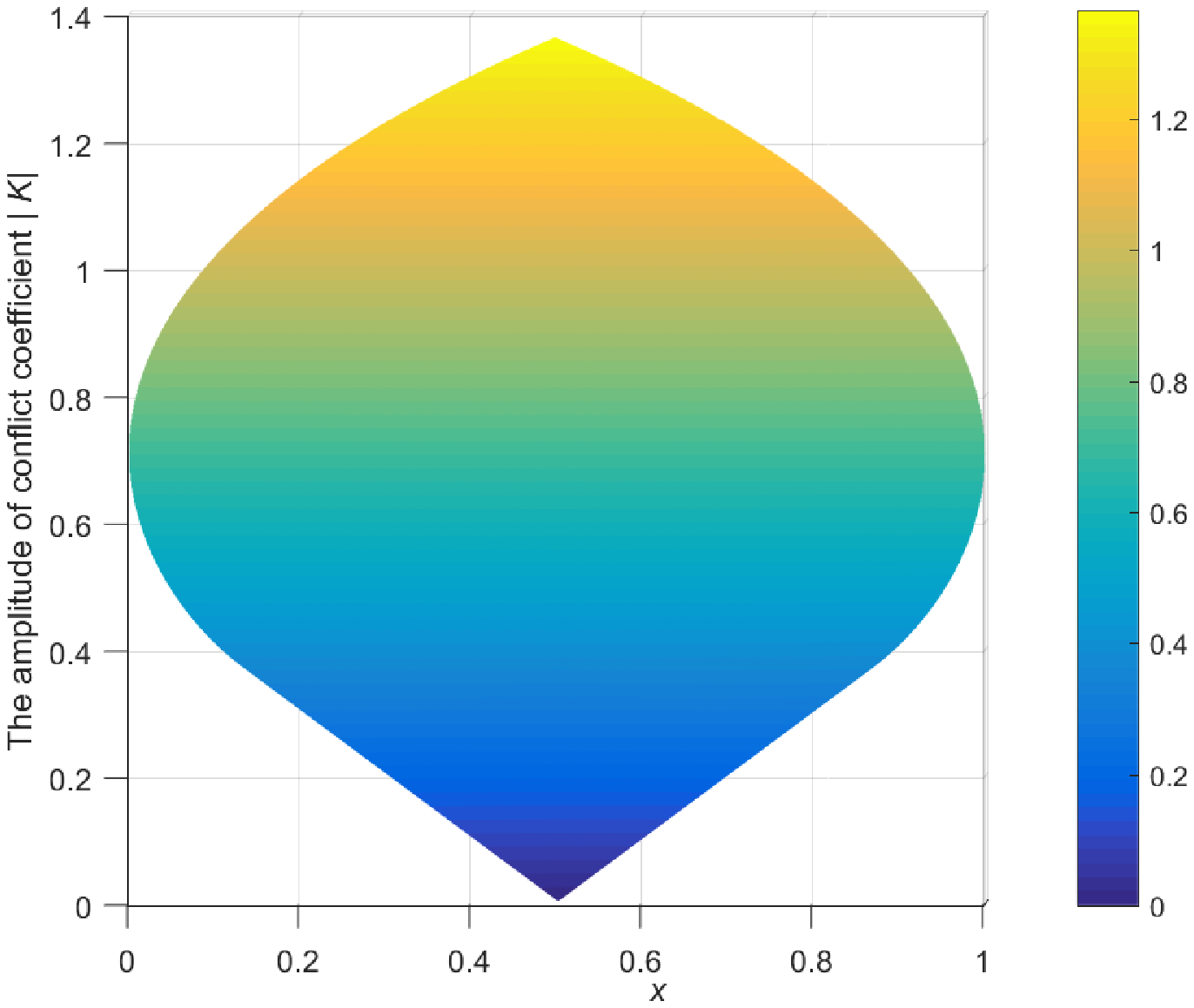}
}
\subfigure[]{
\label{conflictcoefficient:b} 
\centering
\includegraphics[width=.45\linewidth,clip]{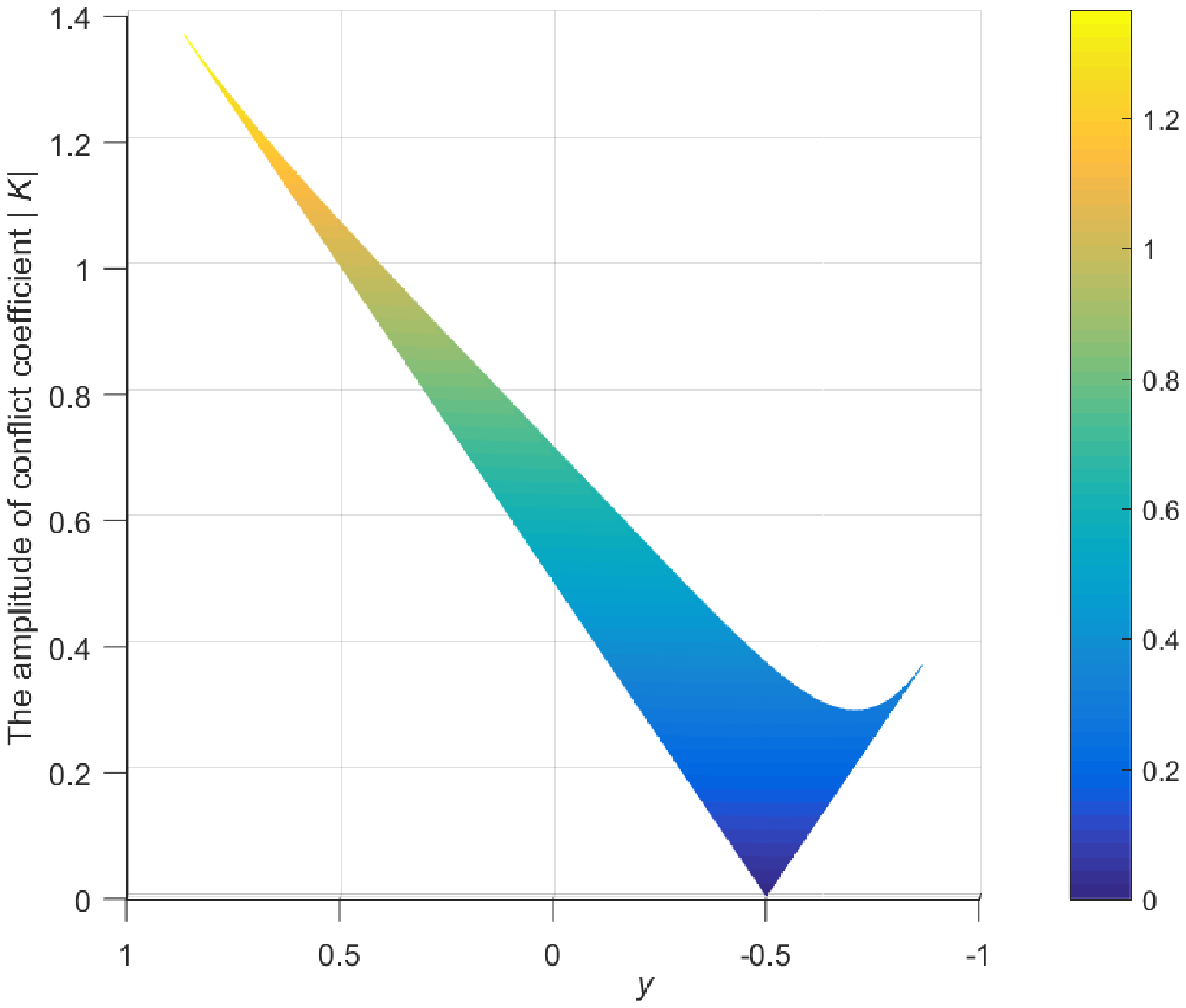}
}
\caption{An example of the variation of $|\mathds{K}|$ between two CBBAs from the front and the side angles.}
\label{conflictcoefficient12} 
\end{figure*}

\begin{figure*}[htbp]
\centering
\subfigure[]{
\label{conflictcoefficient:a} 
\includegraphics[width=.45\linewidth,clip]{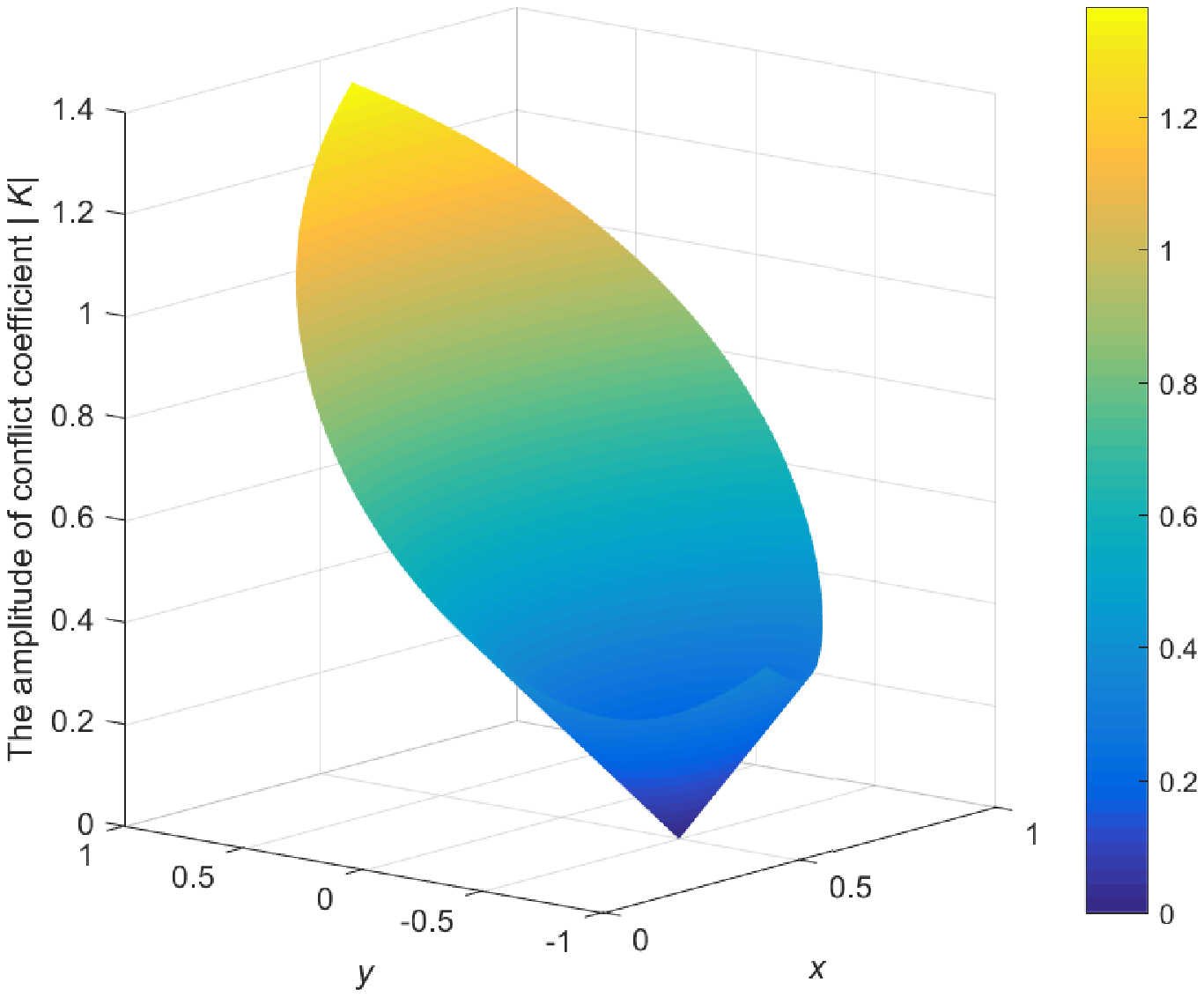}
}
\subfigure[]{
\label{conflictcoefficient:b} 
\centering
\includegraphics[width=.45\linewidth,clip]{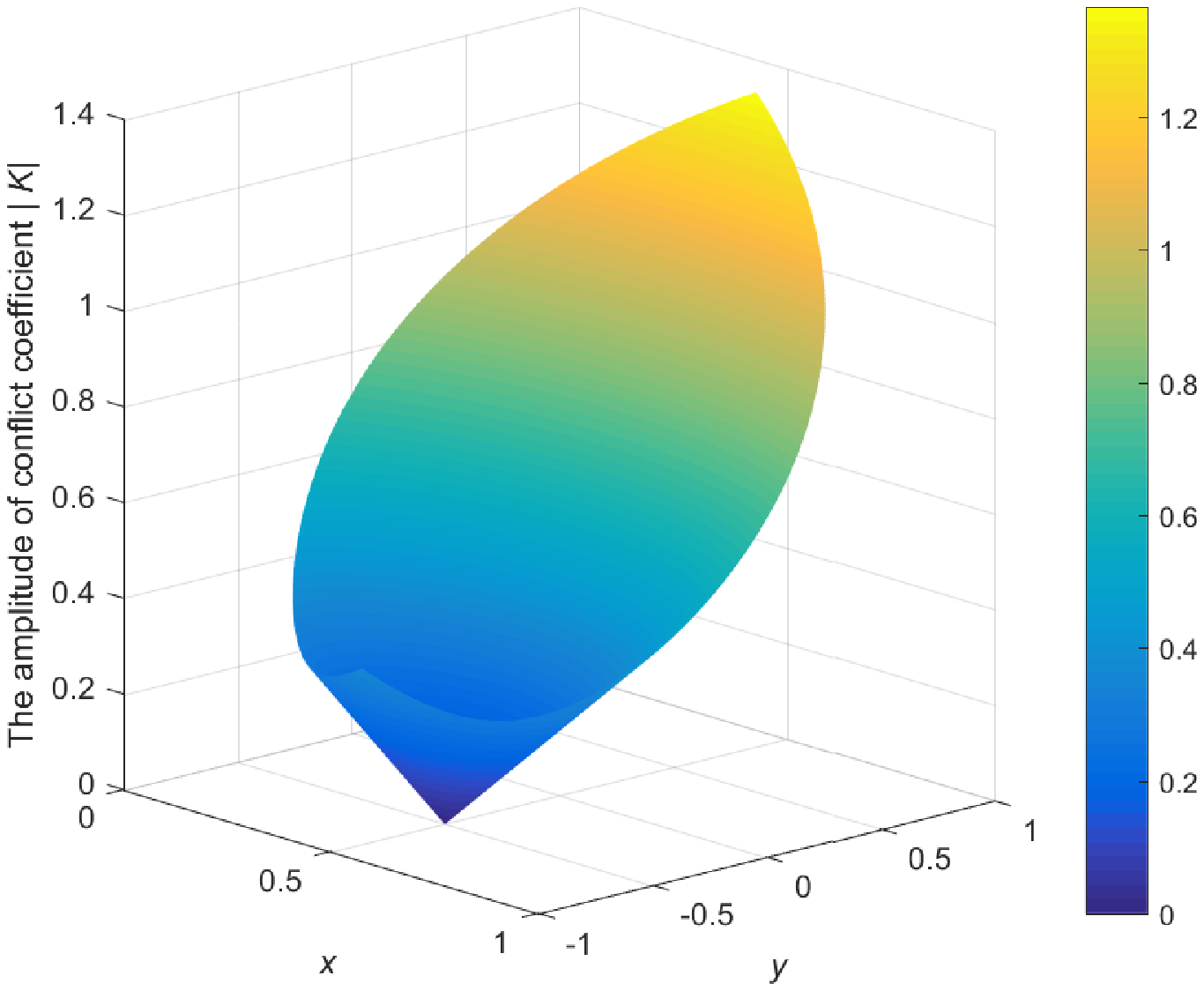}
}
\caption{An example of the variation of $|\mathds{K}|$ between two CBBAs from the oblique angles.}
\label{conflictcoefficient34} 
\end{figure*}

Fig.~\ref{conflictcoefficient12} and Fig.~\ref{conflictcoefficient34} show the results of the magnitude of conflict coefficient $|\mathds{K}|$ between the two CBBAs $\mathds{M}_1$ and $\mathds{M}_2$ from different angles.

In particular, as shown in Fig.~\ref{conflictcoefficient12}, in the case that $x=1$ and $y=0$, we can obtain that
\begin{align*}
&\mathds{M}_1(A)=1 e^{i \arctan (0)}=1, \\
&\mathds{M}_1(B)=0 e^{i \arctan (0)}=0.
\end{align*}

The conflict coefficient $\mathds{K}$ is calculated as
\begin{align*}
\mathds{K} = 1 \times \sqrt{0.5} e^{i \arctan (-1)} + 0 \times \sqrt{0.5} e^{i \arctan (1)}.
\end{align*}

Then, the magnitude of conflict coefficient $|\mathds{K}|$ between the two CBBAs $m_1$ and $m_2$ is 0.7071.

When $x=0$ and $y=0$, it is obtained that
\begin{align*}
\mathds{M}_1(A)&=0 e^{i \arctan (0)}=0, \\
\mathds{M}_1(B)&=1 e^{i \arctan (0)}=1.
\end{align*}

The conflict coefficient $\mathds{K}$ is calculated as
\begin{align*}
\mathds{K} = 0 \times \sqrt{0.5} e^{i \arctan (-1)} + 1 \times \sqrt{0.5} e^{i \arctan (1)}.
\end{align*}

Then, the magnitude of conflict coefficient  $|\mathds{K}|$ between the two CBBAs $\mathds{M}_1$ and $\mathds{M}_2$ is calculated by
\begin{align*}
|\mathds{K}| = 0.7071,
\end{align*}
which shows the same result as the case that $x=1$ and $y=0$.

In the case that $x=0.5$ and $y=-0.8660$, we can obtain that
\begin{align*}
\mathds{M}_1(A)=1 e^{i \arctan (\frac{-0.8660}{0.5})}=1 e^{i \arctan (-1.7320)}, \\
\mathds{M}_1(B)=1 e^{i \arctan (\frac{0.8660}{0.5})}=1 e^{i \arctan (1.7320)}.
\end{align*}

The conflict coefficient $\mathds{K}$ is calculated as
\begin{align*}
\mathds{K}=
&1 e^{i \arctan (-1.7320)} \times \sqrt{0.5} e^{i \arctan (-1)} + \\
&1 e^{i \arctan (1.7320)} \times \sqrt{0.5} e^{i \arctan (1)}.
\end{align*}

Then, the magnitude of conflict coefficient $|\mathds{K}|$ between the two CBBAs $\mathds{M}_1$ and $\mathds{M}_2$ is calculated by
\begin{align*}
|\mathds{K}| = 0.3660.
\end{align*}

In the case that $x=0.5$ and $y=-0.5$, we can obtain that
\begin{align*}
\mathds{M}_1(A)&=\sqrt{0.5} e^{i \arctan (-1)}, \\
\mathds{M}_1(B)&=\sqrt{0.5} e^{i \arctan (1)}.
\end{align*}

The conflict coefficient $\mathds{K}$ is calculated as
\begin{align*}
\mathds{K}=
&\sqrt{0.5} e^{i \arctan (-1)} \times \sqrt{0.5} e^{i \arctan (-1)} + \\
&\sqrt{0.5} e^{i \arctan (1)} \times \sqrt{0.5} e^{i \arctan (1)}.
\end{align*}

Then, the magnitude of conflict coefficient $|\mathds{K}|$ between the two CBBAs $\mathds{M}_1$ and $\mathds{M}_2$ is calculated by
\begin{align*}
|\mathds{K}| = 0.
\end{align*}

When $x=0.5$ and $y=0.8660$, it is obtained that
\begin{align*}
\mathds{M}_1(A)&=1 e^{i \arctan (\frac{0.8660}{0.5})}=1 e^{i \arctan (1.7320)}, \\
\mathds{M}_1(B)&=1 e^{i \arctan (\frac{-0.8660}{0.5})}=1 e^{i \arctan (-1.7320)}.
\end{align*}

The conflict coefficient $\mathds{K}$ is calculated as
\begin{align*}
\mathds{K}=
&1 e^{i \arctan (1.7320)} \times \sqrt{0.5} e^{i \arctan (-1)} + \\
&1 e^{i \arctan (-1.7320)} \times \sqrt{0.5} e^{i \arctan (1)}.
\end{align*}

Then, the magnitude of conflict coefficient  $|\mathds{K}|$ between the two CBBAs $\mathds{M}_1$ and $\mathds{M}_2$ is calculated by
\begin{align*}
|\mathds{K}| = 1.3660.
\end{align*}

\section{Numerical examples}\label{Experiments}
In this section, several numerical examples are illustrated to show the effectiveness of the generalized Dempster--Shafer evidence theory.

\begin{exmp}\label{exa_1}
\rm Supposing that there are two CBBAs $\mathds{M}_1$ and $\mathds{M}_2$ in the frame of discernment $\Omega=\{A,B\}$, and the two CBBAs are given as follows:
\end{exmp}

\begin{align*}
\mathds{M}_1:
&\mathds{M}_1(A)=0.2031 e^{i \arctan(-1.7678)}, \\
&\mathds{M}_1(B)=0.7842 e^{i \arctan(0.5051)}, \\
&\mathds{M}_1(A,B)=0.2669 e^{i \arctan(-0.8839)};\\
\mathds{M}_2:
&\mathds{M}_2(A)=0.3606 e^{i \arctan(3.4641)}, \\
&\mathds{M}_2(B)=0.6245 e^{i \arctan(0.2887)}, \\
&\mathds{M}_2(A,B)=0.6000 e^{i \arctan(-1.7321)}.
\end{align*}

Then, the fusing results are calculated by utilising Eq.~(\ref{eq_GDempsterrule1}) as follows:

\begin{tabular}[t]{l}
$\mathds{M}(A) = 0.0997 e^{i \arctan(0.1900)} = 0.0979 + 0.0186i$, \\
$\mathds{M}(B) = 0.9213 e^{i \arctan(-0.2015)} = 0.9031 - 0.1820i$, \\
$\mathds{M}(A,B) = 0.1634 e^{i \arctan(-163.4)} = -0.0010 + 0.1634i$. \\
\specialrule{0em}{6pt}{6pt}
\end{tabular}

It is verified that $\mathds{M}(A)$ + $\mathds{M}(B)$ + $\mathds{M}(A,B)$ = 1 in this example.

\begin{exmp}\label{exa_2}
\rm Supposing that there are two CBBAs $\mathds{M}_1$ and $\mathds{M}_2$ in the frame of discernment $\Omega=\{A,B\}$, and the two CBBAs are given as follows:
\end{exmp}

\begin{align*}
\mathds{M}_1:
&\mathds{M}_1(A)=0.3606 e^{i \arctan(3.4641)}, \\
&\mathds{M}_1(B)=0.6245 e^{i \arctan(0.2887)}, \\
&\mathds{M}_1(A,B)=0.6000 e^{i \arctan(-1.7321)};
\end{align*}
\begin{align*}
\mathds{M}_2:
&\mathds{M}_2(A)=0.2031 e^{i \arctan(-1.7678)}, \\
&\mathds{M}_2(B)=0.7842 e^{i \arctan(0.5051)}, \\
&\mathds{M}_2(A,B)=0.2669 e^{i \arctan(-0.8839)}.
\end{align*}

The fusing results by utilising Eq.~(\ref{eq_GDempsterrule1}) are calculated as follows:

\begin{tabular}[t]{l}
$\mathds{M}(A) = 0.0997 e^{i \arctan(0.1900)} = 0.0979 + 0.0186i$, \\
$\mathds{M}(B) = 0.9213 e^{i \arctan(-0.2015)} = 0.9031 - 0.1820i$, \\
$\mathds{M}(A,B) = 0.1634 e^{i \arctan(-163.4)} = -0.0010 + 0.1634i$. \\
\specialrule{0em}{6pt}{6pt}
\end{tabular}

It is obvious that $\mathds{M}(A)$ + $\mathds{M}(B)$ + $\mathds{M}(A,B)$ = 1 in this example.

Through Example~\ref{exa_1} and Example~\ref{exa_2}, it verifies that the generalized Dempster--Shafer evidence theory satisfies the commutative law.

\begin{exmp}\label{exa_3}
\rm Supposing that there are two CBBAs $\mathds{M}_1$ and $\mathds{M}_2$ in the frame of discernment $\Omega=\{A,B\}$ where they are degenerated to real numbers, and the two CBBAs are given as follows:
\end{exmp}

\begin{tabular}[t]{l}
$\mathds{M}_1:$
$\mathds{M}_1(A)=0.8$,
$\mathds{M}_1(B)=0.2$;\\
$\mathds{M}_2:$
$\mathds{M}_2(A)=0.9$,
$\mathds{M}_2(B)=0.1$.\\
\specialrule{0em}{6pt}{6pt}
\end{tabular}

On the one hand, by utilising Eq.~(\ref{eq_GDempsterrule1}) of the generalized Dempster's rule of combination, the fusing results are generated as follows:

\begin{tabular}[t]{l}
$\mathds{M}(A)=0.9730$,\\
$\mathds{M}(B)=0.0270$;\\
\specialrule{0em}{6pt}{6pt}
\end{tabular}

On the other hand, based on Eq.~(\ref{eq_Dempsterrule1}) of the classical Dempster's rule of combination, the fusing results are calculated as follows:

\begin{tabular}[t]{l}
$\mathds{M}(A)=0.9730$,\\
$\mathds{M}(B)=0.0270$;\\
\specialrule{0em}{6pt}{6pt}
\end{tabular}

It is easy to see that the fusing results from the generalized Dempster's rule of combination is exactly the same as the fusing results from the classical Dempster's rule of combination.
In this example, the conflict coefficient $\mathds{K}$ is 0.2600.

This example verifies that when the complex mass function is degenerated from complex numbers to real numbers, the generalized Dempster's combination rule degenerates to the classical evidence theory under the condition that the conflict coefficient between the evidences $\mathds{K}$ is less than 1.

\section{Conclusions}\label{Conclusion}
In this paper, a generalized Dempster--Shafer (GDS) evidence theory is proposed.
The main contribution of this study is that a mass function in the GDS evidence theory is modeled as a complex number, called as a complex basic belief assignment.
In addition, the definitions of complex belief function and complex plausibility function are also presented in this paper.
Based on that, a generalized Dempster's rule of combination is exploited to fuse the complex basic belief assignments.
When the complex mass function is degenerated from complex numbers to real numbers, the GDS evidence theory degenerates to the traditional evidence theory under the condition that the conflict coefficient between the evidences $\mathds{K}$ is less than 1.
In summary, this study is the first work to generalize the evidence theory in the framework of complex numbers.
It provides a promising way to model and handle more uncertain information in the process of solving the decision-making problems.

\section*{Acknowledgment}
This research is supported by the Fundamental Research Funds for the Central Universities (No. XDJK2019C085) and Chongqing Overseas Scholars Innovation Program (No. cx2018077).

\ifCLASSOPTIONcaptionsoff
  \newpage
\fi

\bibliographystyle{IEEEtran}

\begin{thebibliography}{10}
\providecommand{\url}[1]{#1}
\csname url@samestyle\endcsname
\providecommand{\newblock}{\relax}
\providecommand{\bibinfo}[2]{#2}
\providecommand{\BIBentrySTDinterwordspacing}{\spaceskip=0pt\relax}
\providecommand{\BIBentryALTinterwordstretchfactor}{4}
\providecommand{\BIBentryALTinterwordspacing}{\spaceskip=\fontdimen2\font plus
\BIBentryALTinterwordstretchfactor\fontdimen3\font minus
  \fontdimen4\font\relax}
\providecommand{\BIBforeignlanguage}[2]{{%
\expandafter\ifx\csname l@#1\endcsname\relax
\typeout{** WARNING: IEEEtran.bst: No hyphenation pattern has been}%
\typeout{** loaded for the language `#1'. Using the pattern for}%
\typeout{** the default language instead.}%
\else
\language=\csname l@#1\endcsname
\fi
#2}}
\providecommand{\BIBdecl}{\relax}
\BIBdecl

\bibitem{yager2018using}
R.~R. Yager, ``On using the shapley value to approximate the {Choquet} integral
  in cases of uncertain arguments,'' \emph{IEEE Transactions on Fuzzy Systems},
  vol.~26, no.~3, pp. 1303--1310, 2018.

\bibitem{zavadskas2014extension}
E.~K. Zavadskas, J.~Antucheviciene, S.~H.~R. Hajiagha, and S.~S. Hashemi,
  ``Extension of weighted aggregated sum product assessment with
  interval-valued intuitionistic fuzzy numbers {(WASPAS-IVIF)},'' \emph{Applied
  Soft Computing}, vol.~24, pp. 1013--1021, 2014.

\bibitem{fu2018data}
C.~Fu, W.~Liu, and W.~Chang, ``Data-driven multiple criteria decision making
  for diagnosis of thyroid cancer,'' \emph{Annals of Operations Research}, pp.
  1--30, 2018.

\bibitem{fei2019interval}
L.~Fei, ``On interval-valued fuzzy decision-making using soft likelihood
  functions,'' \emph{International Journal of Intelligent Systems}, 2019.

\bibitem{Xiao2019Divergence}
F.~Xiao and W.~Ding, ``Divergence measure of {Pythagorean} fuzzy sets and its
  application in medical diagnosis,'' \emph{Applied Soft Computing}, vol.~79,
  pp. 254--267, 2019.

\bibitem{feng2016soft}
F.~Feng, J.~Cho, W.~Pedrycz, H.~Fujita, and T.~Herawan, ``Soft set based
  association rule mining,'' \emph{Knowledge-Based Systems}, vol. 111, pp.
  268--282, 2016.

\bibitem{Sun2019A}
R.~Sun and Y.~Deng, ``{A new method to identify incomplete frame of discernment
  in evidence theory},'' \emph{IEEE Access}, vol.~7, no.~1, pp.
  15\,547--15\,555, 2019.

\bibitem{jiang2018Correlation}
W.~Jiang, ``A correlation coefficient for belief functions,''
  \emph{International Journal of Approximate Reasoning}, vol. 103, pp. 94--106,
  2018.

\bibitem{Zhao2019Dnumbers}
J.~Zhao and Y.~Deng, ``{Performer selection in Human Reliability analysis: D
  numbers approach},'' \emph{International Journal of Computers Communications
  \& Control}, vol.~14, no.~4, pp. 521--536, 2019.

\bibitem{DNTIJAR2019}
X.~Deng and W.~Jiang, ``D number theory based game-theoretic framework in
  adversarial decision making under a fuzzy environment,'' \emph{International
  Journal of Approximate Reasoning}, vol. 106, pp. 194--213, 2019.

\bibitem{Jiang2019Znetwork}
W.~Jiang, Y.~Cao, and X.~Deng, ``{A Novel Z-network Model Based on Bayesian
  Network and Z-number},'' \emph{IEEE Transactions on Fuzzy Systems}, 2019.

\bibitem{kang2019environmental}
B.~Kang, P.~Zhang, Z.~Gao, G.~Chhipi-Shrestha, K.~Hewage, and R.~Sadiq,
  ``Environmental assessment under uncertainty using {Dempster--Shafer} theory
  and {Z-numbers},'' \emph{Journal of Ambient Intelligence and Humanized
  Computing}, pp. DOI: 10.1007/s12\,652--019--01\,228--y, 2019.

\bibitem{Seiti2019Rnumbers}
H.~Seiti, A.~Hafezalkotob, and L.~Mart{\'\i}nez, ``R-numbers, a new risk
  modeling associated with fuzzy numbers and its application to decision
  making,'' \emph{Information Sciences}, vol. 483, pp. 206--231, 2019.

\bibitem{seiti2019developing}
H.~Seiti and A.~Hafezalkotob, ``Developing the {R-TOPSIS} methodology for
  risk-based preventive maintenance planning: A case study in rolling mill
  company,'' \emph{Computers \& Industrial Engineering}, vol. 128, pp.
  622--636, 2019.

\bibitem{cao2018Inherent}
Z.~Cao and C.~T. Lin, ``Inherent fuzzy entropy for the improvement of {EEG}
  complexity evaluation,'' \emph{IEEE Transactions on Fuzzy Systems}, vol.~26,
  no.~2, pp. 1032--1035, 2018.

\bibitem{wang2018analysis}
Q.~Wang, Y.~Li, and X.~Liu, ``Analysis of feature fatigue {EEG} signals based
  on wavelet entropy,'' \emph{International Journal of Pattern Recognition and
  Artificial Intelligence}, vol.~32, no.~08, p. 1854023, 2018.

\bibitem{yager2019using}
R.~R. Yager and F.~E. Petry, ``Using quality measures in the intelligent fusion
  of probabilistic information,'' in \emph{Information Quality in Information
  Fusion and Decision Making}.\hskip 1em plus 0.5em minus 0.4em\relax Springer,
  2019, pp. 51--77.

\bibitem{seiti2018extending}
H.~Seiti, A.~Hafezalkotob, and R.~Fattahi, ``Extending a
  pessimistic--optimistic fuzzy information axiom based approach considering
  acceptable risk: Application in the selection of maintenance strategy,''
  \emph{Applied Soft Computing}, vol.~67, pp. 895--909, 2018.

\bibitem{deng2019evaluating}
X.~Deng and W.~Jiang, ``Evaluating green supply chain management practices
  under fuzzy environment: a novel method based on {D} number theory,''
  \emph{International Journal of Fuzzy Systems}, pp. DOI:
  10.1007/s40\,815--019--00\,639--5, 2019.

\bibitem{Geng2019Saliency}
J.~Geng, X.~Ma, X.~Zhou, and H.~Wang, ``Saliency-guided deep neural networks
  for {SAR} image change detection,'' \emph{IEEE Transactions on Geoscience and
  Remote Sensing}, pp. 1--13, 2019.

\bibitem{zhou2019robust}
D.~Zhou, A.~Al-Durra, K.~Zhang, A.~Ravey, and F.~Gao, ``A robust prognostic
  indicator for renewable energy technologies: A novel error correction grey
  prediction model,'' \emph{IEEE Transactions on Industrial Electronics}, p.
  DOI: 10.1109/TIE.2019.2893867, 2019.

\bibitem{cao2019extraction}
Z.~Cao, C.-T. Lin, K.-L. Lai, L.-W. Ko, J.-T. King, K.-K. Liao, J.-L. Fuh, and
  S.-J. Wang, ``Extraction of {SSVEPs-based} inherent fuzzy entropy using a
  wearable headband {EEG} in migraine patients,'' \emph{IEEE Transactions on
  Fuzzy Systems}, p. DOI: 10.1109/TFUZZ.2019.2905823, 2019.

\bibitem{Xiao2018Anovelmulti}
F.~Xiao, ``A novel multi-criteria decision making method for assessing
  health-care waste treatment technologies based on {D} numbers,''
  \emph{Engineering Applications of Artificial Intelligence}, vol.~71, no.
  2018, pp. 216--225, 2018.

\bibitem{feng2018another}
F.~Feng, H.~Fujita, M.~I. Ali, R.~R. Yager, and X.~Liu, ``Another view on
  generalized intuitionistic fuzzy soft sets and related multiattribute
  decision making methods,'' \emph{IEEE Transactions on Fuzzy Systems},
  vol.~27, no.~3, pp. 474--488, 2018.

\bibitem{Xiao2019Amultiplecriteria}
F.~Xiao, ``A multiple criteria decision-making method based on {D} numbers and
  belief entropy,'' \emph{International Journal of Fuzzy Systems}, vol.~21,
  no.~4, pp. 1144--1153, 2019.

\bibitem{Dempster1967Upper}
A.~P. Dempster, ``Upper and lower probabilities induced by a multivalued
  mapping,'' \emph{Annals of Mathematical Statistics}, vol.~38, no.~2, pp.
  325--339, 1967.

\bibitem{shafer1976mathematical}
G.~Shafer \emph{et~al.}, \emph{A mathematical theory of evidence}.\hskip 1em
  plus 0.5em minus 0.4em\relax Princeton University Press Princeton, 1976,
  vol.~1.

\bibitem{XDeng2019Polymatrix}
X.~Deng, W.~Jiang, and Z.~Wang, ``Zero-sum polymatrix games with link
  uncertainty: {A} {D}empster-{S}hafer theory solution,'' \emph{Applied
  Mathematics and Computation}, vol. 340, pp. 101--112, 2019.

\bibitem{suxiaoyan2019}
X.~Su, L.~Li, H.~Qian, M.~Sankaran, and Y.~Deng, ``A new rule to combine
  dependent bodies of evidence,'' \emph{Soft Computing}, pp. DOI:
  10.1007/s00\,500--019--03\,804--y, 2019.

\bibitem{yager2017soft}
R.~R. Yager, P.~Elmore, and F.~Petry, ``Soft likelihood functions in combining
  evidence,'' \emph{Information Fusion}, vol.~36, pp. 185--190, 2017.

\bibitem{lilusu2018}
X.~Su, L.~Li, F.~Shi, and H.~Qian, ``Research on the fusion of dependent
  evidence based on mutual information,'' \emph{IEEE Access}, vol.~6, pp.
  71\,839--71\,845, 2018.

\bibitem{yager2018satisfying}
R.~R. Yager, ``Satisfying uncertain targets using measure generalized
  {Dempster-Shafer} belief structures,'' \emph{Knowledge-Based Systems}, vol.
  142, pp. 1--6, 2018.

\bibitem{seiti2018risk}
H.~Seiti, A.~Hafezalkotob, S.~Najafi, and M.~Khalaj, ``A risk-based fuzzy
  evidential framework for {FMEA} analysis under uncertainty: An
  interval-valued {DS} approach,'' \emph{Journal of Intelligent \& Fuzzy
  Systems}, no. Preprint, pp. 1--12, 2018.

\bibitem{Xiao2018AHybridFuzzy}
F.~Xiao, ``A hybrid fuzzy soft sets decision making method in medical
  diagnosis,'' \emph{IEEE Access}, vol.~6, pp. 25\,300--25\,312, 2018.

\bibitem{yager2019generalized}
R.~R. Yager, ``Generalized {Dempster--Shafer} structures,'' \emph{IEEE
  Transactions on Fuzzy Systems}, vol.~27, no.~3, pp. 428--435, 2019.

\bibitem{Jiang2018Information}
Z.~He and W.~Jiang, ``{An evidential Markov decision making model},''
  \emph{Information Sciences}, vol. 467, pp. 357--372, 2018.

\bibitem{yager2018fuzzy}
R.~R. Yager, ``Fuzzy rule bases with generalized belief structure inputs,''
  \emph{Engineering Applications of Artificial Intelligence}, vol.~72, pp.
  93--98, 2018.

\bibitem{ablowitz2003complex}
M.~J. Ablowitz and A.~S. Fokas, \emph{Complex variables: introduction and
  applications}.\hskip 1em plus 0.5em minus 0.4em\relax Cambridge University
  Press, 2003.

\bibitem{zavadskas2017model}
E.~K. Zavadskas, R.~Bausys, B.~Juodagalviene, and I.~Garnyte-Sapranaviciene,
  ``Model for residential house element and material selection by neutrosophic
  {MULTIMOORA} method,'' \emph{Engineering Applications of Artificial
  Intelligence}, vol.~64, pp. 315--324, 2017.

\bibitem{zhou2018evidential}
M.~Zhou, X.-B. Liu, Y.-W. Chen, and J.-B. Yang, ``Evidential reasoning rule for
  {MADM} with both weights and reliabilities in group decision making,''
  \emph{Knowledge-Based Systems}, vol. 143, pp. 142--161, 2018.

\bibitem{de2018robust}
V.~H.~C. de~Albuquerque, T.~M. Nunes, D.~R. Pereira, E.~J. d.~S. Luz,
  D.~Menotti, J.~P. Papa, and J.~M.~R. Tavares, ``Robust automated cardiac
  arrhythmia detection in {ECG} beat signals,'' \emph{Neural Computing and
  Applications}, vol.~29, no.~3, pp. 679--693, 2018.

\bibitem{yager2018multi}
R.~R. Yager, ``Multi-criteria decision making with interval criteria
  satisfactions using the golden rule representative value,'' \emph{IEEE
  Transactions on Fuzzy Systems}, vol.~26, no.~2, pp. 1023--1031, 2018.

\bibitem{feng2019lexicographic}
F.~Feng, M.~Liang, H.~Fujita, R.~R. Yager, and X.~Liu, ``Lexicographic orders
  of intuitionistic fuzzy values and their relationships,'' \emph{Mathematics},
  vol.~7, no.~2, pp. 1--26, 2019.

\bibitem{Wang2018uncertainty}
X.~Wang and Y.~Song, ``Uncertainty measure in evidence theory with its
  applications,'' \emph{Applied Intelligence}, vol.~48, no.~7, pp. 1672--1688,
  2018.

\bibitem{Li2018Generalized}
Y.~Li and Y.~Deng, ``Generalized ordered propositions fusion based on belief
  entropy,'' \emph{International Journal of Computers Communications \&
  Control}, vol.~13, no.~5, pp. 792--807, 2018.

\bibitem{liu2018classifier}
Z.~Liu, Q.~Pan, J.~Dezert, J.-W. Han, and Y.~He, ``Classifier fusion with
  contextual reliability evaluation,'' \emph{IEEE Transactions on Cybernetics},
  vol.~48, no.~5, pp. 1605--1618, 2018.

\bibitem{gong2018Research}
Y.~Gong, X.~Su, H.~Qian, and N.~Yang, ``Research on fault diagnosis methods for
  the reactor coolant system of nuclear power plant based on {D}-{S} evidence
  theory,'' \emph{Annals of Nuclear Energy}, pp. 395--399, 2018.

\bibitem{Jiang2019IJIS}
W.~Jiang, C.~Huang, and X.~Deng, ``A new probability transformation method
  based on a correlation coefficient of belief functions,'' \emph{International
  Journal of Intelligent Systems}, pp. In press, DOI: 10.1002/int.22\,098,
  2019.

\bibitem{Xiao2019Multisensor}
F.~Xiao, ``Multi-sensor data fusion based on the belief divergence measure of
  evidences and the belief entropy,'' \emph{Information Fusion}, vol.~46, no.
  2019, pp. 23--32, 2019.

\bibitem{Zhang2018DEMATEL}
W.~Zhang and Y.~Deng, ``{Combining conflicting evidence using the DEMATEL
  method},'' \emph{Soft Computing}, pp. DOI: 10.1007/s00\,500--018--3455--8,
  2018.

\bibitem{Sun2019GBPA}
R.~Sun and Y.~Deng, ``A new method to determine generalized basic probability
  assignment in the open world,'' \emph{IEEE Access}, vol.~7, no.~1, pp.
  52\,827--52\,835, 2019.

\bibitem{dezert2018total}
J.~Dezert, A.~Tchamova, and D.~Han, ``Total belief theorem and conditional
  belief functions,'' \emph{International Journal of Intelligent Systems},
  vol.~33, no.~12, pp. 2314--2340, 2018.

\bibitem{Jiang2018KBS}
Z.~He and W.~Jiang, ``An evidential dynamical model to predict the interference
  effect of categorization on decision making results,'' \emph{Knowledge-Based
  Systems}, vol. 150, pp. 139--149, 2018.

\bibitem{Li2019TDBF}
Y.~Li and Y.~Deng, ``{TDBF: Two Dimension Belief Function},''
  \emph{International Journal of Intelligent Systems}, vol.~34, p. DOI:
  10.1002/int.22135, 2019.

\bibitem{yager2019entailment}
R.~R. Yager, ``Entailment for measure based belief structures,''
  \emph{Information Fusion}, vol.~47, pp. 111--116, 2019.

\bibitem{Gao2019negation}
X.~Gao and Y.~Deng, ``The negation of basic probability assignment,''
  \emph{IEEE Access}, vol.~7, no.~1, p. DOI: 10.1109/ACCESS.2019.2901932, 2019.

\bibitem{Gao2019generalizationnegation}
------, ``The generalization negation of probability distribution and its
  application in target recognition based on sensor fusion,''
  \emph{International Journal of Distributed Sensor Networks}, vol.~15, no.~5,
  p. DOI: 10.1177/1550147719849381, 2019.

\bibitem{han2016belief}
D.~Han, J.~Dezert, and Y.~Yang, ``Belief interval-based distance measures in
  the theory of belief functions,'' \emph{IEEE Transactions on Systems, Man,
  and Cybernetics: Systems}, vol.~48, no.~6, pp. 833--850, 2016.

\bibitem{Songyf2016}
Y.~Song, X.~Wang, L.~Lei, and S.~Yue, ``{Uncertainty measure for
  interval-valued belief structures},'' \emph{{Measurement}}, vol.~{80}, pp.
  {241--250}, {2016}.

\bibitem{Song2019divergence}
Y.~Song and Y.~Deng, ``A new method to measure the divergence in evidential
  sensor data fusion,'' \emph{International Journal of Distributed Sensor
  Networks}, vol.~15, no.~4, p. DOI: 10.1177/1550147719841295, 2019.

\bibitem{cui2019improved}
H.~Cui, Q.~Liu, J.~Zhang, and B.~Kang, ``An improved deng entropy and its
  application in pattern recognition,'' \emph{IEEE Access}, vol.~7, pp.
  18\,284--18\,292, 2019.

\bibitem{yager2008entropy}
R.~R. Yager, ``Entropy and specificity in a mathematical theory of evidence,''
  in \emph{Classic Works of the Dempster-Shafer Theory of Belief
  Functions}.\hskip 1em plus 0.5em minus 0.4em\relax Springer, 2008, pp.
  291--310.

\bibitem{dong2019combination}
Y.~Dong, J.~Zhang, Z.~Li, Y.~Hu, and Y.~Deng, ``Combination of evidential
  sensor reports with distance function and belief entropy in fault
  diagnosis,'' \emph{International Journal of Computers Communications \&
  Control}, vol.~14, no.~3, pp. 293--307, 2019.

\bibitem{liu2018combination}
Z.-G. Liu, Q.~Pan, J.~Dezert, and A.~Martin, ``Combination of classifiers with
  optimal weight based on evidential reasoning,'' \emph{IEEE Transactions on
  Fuzzy Systems}, vol.~26, no.~3, pp. 1217--1230, 2018.

\bibitem{Zhanghp2018AIME}
H.~Zhang and Y.~Deng, ``{ Engine fault diagnosis based on sensor data fusion
  considering information quality and evidence theory},'' \emph{{Advances in
  Mechanical Engineering}}, vol.~{10}, no.~{11}, pp. {1--10}, {2018}.

\bibitem{yager2017maxitive}
R.~R. Yager and N.~Alajlan, ``Maxitive belief structures and imprecise
  possibility distributions,'' \emph{IEEE Transactions on Fuzzy Systems},
  vol.~25, no.~4, pp. 768--774, 2017.

\bibitem{Xu2019DEMATEL}
H.~Xu and Y.~Deng, ``{Dependent Evidence Combination Based on DEMATEL
  Method},'' \emph{International Journal of Intelligent Systems}, p. DOI:
  10.1002/int.22107, 2019.

\bibitem{Jiang2019Uncertainty}
Z.~Huang, L.~Yang, and W.~Jiang, ``Uncertainty measurement with belief entropy
  on the interference effect in the quantum-like {B}ayesian {N}etworks,''
  \emph{Applied Mathematics and Computation}, vol. 347, pp. 417--428, 2019.

\bibitem{zhou2019evidential}
M.~Zhou, X.-B. Liu, J.-B. Yang, Y.-W. Chen, and J.~Wu, ``Evidential reasoning
  approach with multiple kinds of attributes and entropy-based weight
  assignment,'' \emph{Knowledge-Based Systems}, vol. 163, pp. 358--375, 2019.

\bibitem{fu2019multiple}
C.~Fu, W.~Chang, M.~Xue, and S.~Yang, ``Multiple criteria group decision making
  with belief distributions and distributed preference relations,''
  \emph{European Journal of Operational Research}, vol. 273, no.~2, pp.
  623--633, 2019.

\bibitem{song2018sensor}
Y.~Song, X.~Wang, J.~Zhu, and L.~Lei, ``Sensor dynamic reliability evaluation
  based on evidence theory and intuitionistic fuzzy sets,'' \emph{Applied
  Intelligence}, vol.~48, no.~11, pp. 3950--3962, 2018.

\bibitem{fan2018evidence}
C.-l. Fan, Y.~Song, L.~Lei, X.~Wang, and S.~Bai, ``Evidence reasoning for
  temporal uncertain information based on relative reliability evaluation,''
  \emph{Expert Systems with Applications}, vol. 113, pp. 264--276, 2018.

\bibitem{Zhanghp2019Weighted}
H.~Zhang and Y.~Deng, ``Weighted belief function of sensor data fusion in
  engine fault diagnosis,'' \emph{Soft Computing}, pp. DOI:
  10.1007/s00\,500--019--04\,063--7, 2019.

\bibitem{zhou2017evidential}
M.~Zhou, X.~Liu, and J.~Yang, ``Evidential reasoning approach for {MADM} based
  on incomplete interval value,'' \emph{Journal of Intelligent \& Fuzzy
  Systems}, vol.~33, no.~6, pp. 3707--3721, 2017.

\bibitem{dezert2016decision}
J.~Dezert, D.~Han, J.-M. Tacnet, S.~Carladous, and Y.~Yang, ``Decision-making
  with belief interval distance,'' in \emph{International Conference on Belief
  Functions}.\hskip 1em plus 0.5em minus 0.4em\relax Springer, 2016, pp.
  66--74.

\bibitem{liu2019evidence}
Z.~Liu, Y.~Liu, J.~Dezert, and F.~Cuzzolin, ``Evidence combination based on
  credal belief redistribution for pattern classification,'' \emph{IEEE
  Transactions on Fuzzy Systems}, p. DOI: 10.1109/TFUZZ.2019.2911915, 2019.

\bibitem{liu2019new}
Z.-g. Liu, Z.~Zhang, Y.~Liu, J.~Dezert, and Q.~Pan, ``A new pattern
  classification improvement method with local quality matrix based on
  {K-NN},'' \emph{Knowledge-Based Systems}, vol. 164, pp. 336--347, 2019.

\end{thebibliography}
\end{document}